\documentclass[10pt,journal,compsoc]{IEEEtran}

\ifCLASSOPTIONcompsoc
  \usepackage[nocompress]{cite}
\else
  \usepackage{cite}
\fi
\ifCLASSINFOpdf
\else
\fi
\usepackage{bbding}
\usepackage{xspace}
\usepackage{graphicx}
\usepackage{amsmath}
\usepackage{amssymb}
\usepackage{booktabs}
\usepackage{bm}
\usepackage{multirow}
\usepackage{array}
\usepackage{comment}

\usepackage{algorithm}
\usepackage{algpseudocode}
\usepackage{bbm}

\newcommand{\algoName}{\texttt{LocATe}\xspace} 
\newcommand{\tdtal}{3D-TAL\xspace}
\newcommand{\datasetName}{BABEL-TAL-20\xspace}
\newcommand{\btt}{BT20\xspace}

\usepackage[pagebackref,breaklinks,colorlinks]{hyperref}

\usepackage[capitalize]{cleveref}
\crefname{section}{Sec.}{Secs.}
\Crefname{section}{Section}{Sections}
\Crefname{table}{Table}{Tables}
\crefname{table}{Tab.}{Tabs.}

\begin{document}
%
\title{\algoName: End-to-end \underline{Loc}alization of \underline{A}ctions in 3D with \underline{T}ransform\underline{e}rs} %
%
%
%
%
\author{Jiankai Sun,
        Bolei Zhou, 
        Michael J. Black, 
        and~Arjun Chandrasekaran
\IEEEcompsocitemizethanks{\IEEEcompsocthanksitem J. Sun is with Stanford University,
CA, 94305, USA.\\
E-mail: jksun@stanford.edu
\IEEEcompsocthanksitem B. Zhou is with University of California, Los Angeles,
CA, 90095, USA.
\IEEEcompsocthanksitem M. Black and A. Chandrasekaran are with Max Planck Institute for Intelligent Systems, Tubingen, Germany.}
}

\IEEEtitleabstractindextext{%
\begin{abstract}
Understanding a person's behavior from their 3D motion is a fundamental problem in computer vision with many applications. 
An important component of this problem is 3D Temporal Action Localization (3D-TAL), which involves recognizing \emph{what} actions a person is performing, and \emph{when}.
State-of-the-art 3D-TAL methods employ a two-stage approach in which the \emph{action span detection} task and the \emph{action recognition} task are implemented as a cascade.
This approach, however, limits the possibility of error-correction. 
In contrast, we propose LocATe, an end-to-end approach that jointly localizes and recognizes actions in a 3D sequence. 
Further, unlike existing autoregressive models that focus on modeling the local context in a sequence, LocATe's transformer model is capable of capturing long-term correlations between actions in a sequence. 
Unlike transformer-based object-detection and classification models which consider image or patch features as input, the input in 3D-TAL is a long sequence of highly correlated frames. 
To handle the high-dimensional input, we implement an effective input representation, and overcome the diffuse attention across long time horizons by introducing sparse attention in the model. 
LocATe outperforms previous approaches 
on the existing PKU-MMD 3D-TAL benchmark (mAP $= 93.2\%$).
Finally, we argue that benchmark datasets are most useful where there is clear room for performance improvement.
To that end, we introduce a new, challenging, and more realistic benchmark dataset, BABEL-TAL-20 (BT20), where the performance of state-of-the-art methods is significantly worse. 
The dataset and code for the method will be available for research purposes. 
\end{abstract}

}

\maketitle

\IEEEdisplaynontitleabstractindextext

%
\IEEEpeerreviewmaketitle

\begin{figure*}
  \centerline{
  \includegraphics[width=\linewidth]{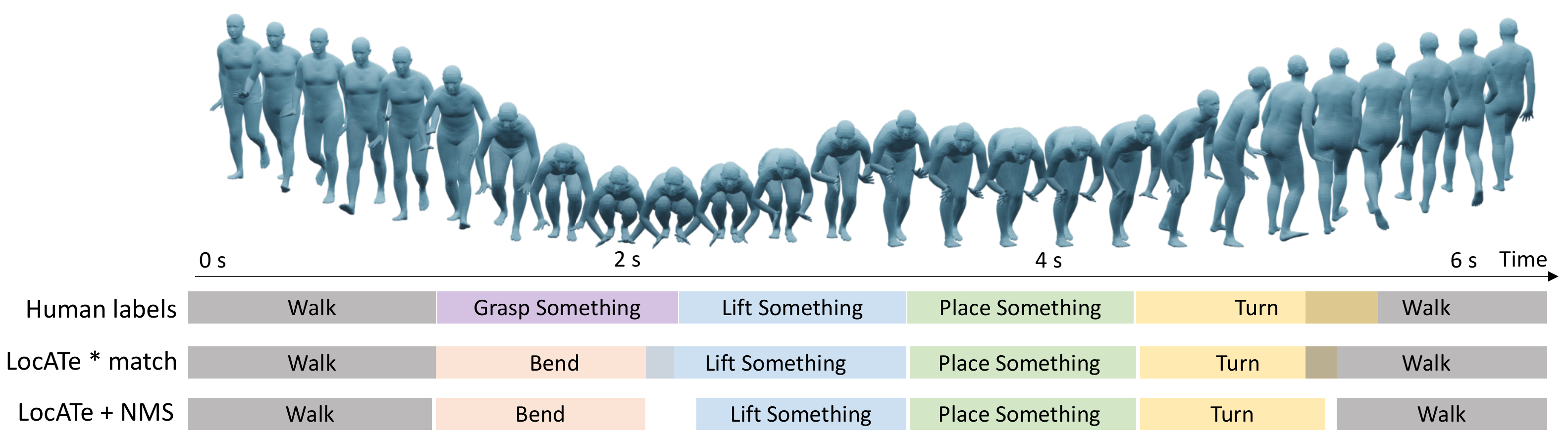}}
  \vspace{-0.1in}
  \caption{
    3D Temporal Action Localization (\tdtal) involves identifying actions and their precise spans (start, end) in a 3D motion. 
    We compare the human-provided labels for the 3D motion (row 1) with predictions from our approach, \algoName (rows 2, 3).
    Row 2 visualizes the action spans from \algoName that best match the human labels, computed via an optimal bi-partite matching (see \cref{sec:method}). 
    Predictions from \algoName correlate well with human labels, including simultaneous actions (visualized as temporal overlap between different spans). 
    For downstream applications, we apply Non-Maximal Suppression (NMS) to the output from \algoName (row 3). 
    \algoName produces accurate localizations, and meaningful actions (even when disagreeing with the human label, e.g., `Grasp Something' vs. `Bend’). 
  } 
  \label{fig:teaser}
\end{figure*}

\section{Introduction}

Understanding a person's behavior in the 3D world is a fundamental and important problem. 
A step towards solving this is identifying what actions a person is performing, when, where, and why. 
3D Temporal Action Localization (\tdtal) involves recognizing the actions that a person is performing in a 3D motion sequence, and locating the precise start and end of each action (see Fig.~\ref{fig:teaser}). 
This has various applications. 
For example, 3D-TAL methods may enable automatic retrieval of semantically relevant movements for graphic artists and game designers who use large databases of 3D data to animate virtual characters. 
In contrast to 2D-TAL methods, 3D-TAL can focus on the body motion alone, factoring out effects of the image texture and lighting, which can bias 2D methods~\cite{choi2019can}. 
Despite the decreasing cost of 3D sensors, e.g., Kinects, phones, and iPads, and the growing accessibility to 3D human movement data \cite{AMASS:ICCV:2019}, progress in 3D-TAL has stagnated in recent years \cite{cui2020skeleton,beyond_joints}, compared to 2D-TAL~\cite{9062498}. 
We argue that the lack of a suitable benchmark dataset limits the rate of progress on a task. 
Current benchmarks are not sufficiently challenging, with saturation in 3D-TAL performance.
Consequently, we propose a new benchmark for 3D-TAL that contains significantly more complex movement sequences, with a more realistic (long-tailed) distribution of actions. 
We use this to evaluate our method, \algoName, which is the first end-end approach for 3D-TAL. \algoName is more accurate, has a simpler architecture, and is faster than existing multi-stage autoregressive methods. 

Current state-of-the-art \tdtal approaches \cite{cui2020skeleton,beyond_joints} are complex, multi-stage pipelines. 
The stages involve prediction of action-agnostic proposals (start, end), 
and performing action recognition on the motion span or on individual frames. 
These approaches have a major shortcoming. Although the localization and recognition tasks are related, they are tackled independently. This eliminates the possibility of error-correction or mechanisms that can mutually adapt the predictions. 
In contrast, the tremendous recent progress in many computer vision tasks has stemmed from deep neural network models that learn representations and solve tasks via end-to-end training on large datasets. 
With this in mind, we present \algoName, an approach that is trained end-to-end and learns a single set of features to jointly recognize and localize actions.
We formulate localization and recognition as an action-span set prediction problem, given the entire input sequence (i.e., $\{$(start time, end time, action), (...) $\}$). 
The set formulation enables \algoName to trivially model simultaneous actions, e.g., `turn' and `walk' in Fig.~\ref{fig:teaser}, unlike previous methods. 

Recently, transformer models have resulted in breakthroughs in natural
language processing~\cite{vaswani2017attention} and computer vision~\cite{dosovitskiy2020image,9712373} due to their ability to
effectively model temporal information over a long time horizon~\cite{qiu2021egocentric,sun2021adversarial}. 
This is critical to the 3D-TAL task since human actions are indeed correlated across time; 
e.g., a person who lifts an object is likely to place it down before picking up another object (e.g., Fig.~\ref{fig:teaser}). 
To model these global correlations effectively, we implement a transformer encoder-decoder architecture \cite{vaswani2017attention} that supports tasks where the input or output can be formulated as a set, e.g., object detection \cite{carion2020end,zhu2020deformable}. 

Unlike other tasks employing transformers in which the inputs are representations of words~\cite{vaswani2017attention}, image patches~\cite{dosovitskiy2020image}, bounding box features~\cite{carion2020end}, etc., the standard input for 3D-TAL consists of a sequence of frames with 3D joint positions. 
While existing methods in 2D-TAL \cite{xu2020g,nawhal2021activity} use features from an action recognition model as the (input) representation for each frame, we find that this is ineffective (see Sup.~Mat.).
Instead, we implement \algoName as a `fully attentional' model, with a sequence of poses (3D joint positions of the human skeleton) as input. 
The model learns to embed short-term pose sequences (8-frame contiguous chunks), and self-attention operations are performed on these short motion chunks (see Sec.~\ref{sec:inp_rep}). 
Despite the decrease in the input feature length of the model due to the input chunks, we observe diffused attention across time-steps. 
To address this issue, we implement a sparse attention mechanism called deformable attention \cite{zhu2020deformable}. 
This use is novel in the context of TAL and significantly improves results (\cref{sec:exp}). 
\algoName has many advantages -- a simpler architecture, faster inference, and superior performance -- compared to previous approaches. 

\algoName achieves mAP $= 93.2\%$ on the current \tdtal~benchmark, PKU-MMD \cite{liu2017pku}. 
While ever-increasing performance on a benchmark \emph{may} indicate progress, we argue that saturating performance on simple benchmark datasets limits progress on a task. 
PKU-MMD is very large in scale, but the subjects were recruited to specifically perform simple everyday actions such as `shake hands', `drink', etc. 
In a proof-of-concept experiment (\cref{tab:pku_mmd}), we observe that \tdtal~approaches can achieve very good performance even if half of the training data is ignored. 
This demonstrates the low intra-class variance among the PKU-MMD motions. 
So it is important to learn from, and evaluate methods on, more complex, long-tailed data, since the frequency of actions in the real world typically follows a long-tailed distribution \cite{kim2020detecting,ramanathan2015learning}. 

The recent BABEL dataset \cite{BABEL:CVPR:2021} contains dense, frame-level action labels for mocap sequences from AMASS \cite{AMASS:ICCV:2019}. 
Actions are labeled post-hoc by human annotators. This results in larger diversity and increases the already-long tail of the action distribution. 
Actions also vary widely in complexity. BABEL contains simple actions like `stand', `walk', `kick', but also complex activities like `dance', `yoga', etc., which have large intra-class variance and are composed of simpler actions. 
We leverage BABEL for the \tdtal~task, and present two experimental settings -- one that is moderately challenging (BT20) and the other extremely challenging (BABEL-TAL-60) (see Sec.~\ref{sec:benchmark}). 
This benchmark presents important new challenges and opportunities for \tdtal~methods: 1) More complex data -- fewer constraints on actions/subjects, and larger intra-class variance. 
2) Multiple labels for actions, reflecting the real-world variability and ambiguity in describing human behavior.
3) Realistic, long-tailed distribution of actions. 

We observe that a previous \tdtal~method \cite{beyond_joints} that achieves mAP $=81.1\%$ on PKU-MMD, achieves only mAP$=11.4\%$ on BT20. This demonstrates the challenging nature of the new benchmark. 
We implement strong baselines by adapting existing 2D-TAL approaches which outperform the previous approach. 
We observe that in addition to outperforming state-of-the-art methods on PKU-MMD, \algoName also achieves state-of-the-art performance on BT20 with mAP $=36.0\%$. 
We further validate the performance improvements via a human study. 
We provide visualizations of the results in the Sup. Mat. 

\noindent
\textbf{Contributions.} 
(1) We implement an end-to-end transformer model that learns to jointly localize and recognize actions. Its architecture is simpler and its predictions are more accurate than previous approaches. 
(2) We observe that while methods approach saturating performance on the current benchmark, the \tdtal task remains unsolved. In an effort to measure meaningful progress in \tdtal, we propose a new benchmark dataset containing more complex, long-tailed actions. 
(3) Previous methods and baselines achieve mAP between $11.4\% - 21.1\%$ on the new benchmark. \algoName outperforms them with mAP $=36.0$. 
This indicates a large room for improvement for methods. 
(4) Trained models, training code, and the dataset will be publicly available for academic research. 

\begin{table*}[t!]
  \centering
  \caption{
    Comparison of large 3D Temporal Action Localization datasets. 
    }
  \begin{tabular}{>{\centering\arraybackslash}m{0.2\linewidth}|>{\centering\arraybackslash}m{0.10\linewidth}|>{\centering\arraybackslash}m{0.10\linewidth}|>{\centering\arraybackslash}m{0.2\linewidth}|>{\centering\arraybackslash}m{0.23\linewidth}}
    \toprule
    Datasets & Classes	& Seq. & Labeled Instances	& Modalities		\\
    \midrule
    \texttt{Comp. Act.}~\cite{6909504} & $16$ & $693$ & $2529$  & RGB+D+Skeleton  \\
    \texttt{Watch-N-Patch}~\cite{7299065} & $21$ & $458$  & $\sim 2500$ & RGB+D+Skeleton  \\
    \texttt{PKU-MMD}~\cite{liu2017pku} & $51$  & $1076$  & $21545$ & RGB+D+IR+Skeleton  \\
    \texttt{BABEL-TAL-60} & $60$  & $8808$  & $70903$  & 3D Mesh+Skeleton \\
    \texttt{\datasetName} & $20$  & $5727$  & $6244$ & 3D Mesh+Skeleton  \\
    \bottomrule
  \end{tabular}
  \label{tab:comp_dataset}
\end{table*}

\section{Related Work}
\label{sec:related_work}

\noindent
\textbf{3D Temporal Action Localization (3D-TAL) datasets.} 
Over the years, datasets have driven progress on \tdtal tasks. 
G3D \cite{bloom2012g3d}, one of the earliest \tdtal datasets, was aimed at understanding the gestures of a person in real-time for video-game applications. 
CAD-120 \cite{Sung2012UnstructuredHA} contains daily activities performed in different environments, but composed of simpler actions. 
Watch-n-Patch \cite{7299065} present a dataset of daily human activities, and an application called `action patching'. 
which involves detecting actions that a person forgets to perform in a long-term activity.  Similarly, 
Lillo et al.~\cite{6909504} focus on modeling spatial and temporal compositions of simple actions to detect complex activities. 
Unlike these datasets, SBU Kinetic interaction \cite{Yun2012TwopersonID} contains $8$ two-person interactions such as `approaching', `hugging', etc. 
PKU-MMD~\cite{liu2017pku} consists of $51$ simple actions such as `putting on glasses', `shake hands', which were specifically performed by actors. 
In contrast, the recent BABEL dataset \cite{BABEL:CVPR:2021} consists of mocap sequences from the large AMASS data archive \cite{AMASS:ICCV:2019}, a consolidation of many mocap datasets. 
BABEL is more complex and has a long-tailed distribution of actions. 
We argue that to truly make progress in \tdtal, it is important to focus on approaches that can efficiently learn from relatively challenging datasets. 

\noindent
\textbf{\tdtal approaches.} 
Prior works have tackled \tdtal using different approaches, but often with the goal of modeling temporal context for better recognition and localization. 
This includes stochastic optimization using Particle Swarm Optimization \cite{papoutsakis2017temporal}, Multiple Instance Learning \cite{Yun2012TwopersonID}, and hierarchical spatio-temporal models  \cite{6909504,7299065,Sung2012UnstructuredHA}. 
Recently, recurrent approaches have been the de-facto standard for modeling temporal context in \tdtal. 
Song et al.~\cite{8322061} utilize spatio-temporal attention to improve recognition accuracy.
Li et al.~\cite{li2016online} propose a joint classification-regression framework.
However, they consider a streaming setting and hence perform per-frame recognition, like Carrera et al.~\cite{carrara2019lstm}. 
Previous state-of-the-art works \cite{beyond_joints,cui2020skeleton} on PKU-MMD treat \tdtal as a per-frame classification, followed by recognition. 
\algoName, however, outputs a \emph{set}, given a motion sequence. 

\noindent
\textbf{2D Temporal Action Localization (2D-TAL).}
2D-TAL is an important, popular task with a rich history~\cite{lin2019bmn,alwassel2021tsp,lin2021learning,bai2020boundary,heilbron2017scc,shou2017cdc,zeng2019graph,tan2021relaxed,9025668,liu2018multi}. 
To localize and recognize actions in untrimmed video sequences, many detection methods utilize a sliding window approach.
Few approaches~\cite{karaman2014fast,oneata2013action,wang2014action} slide the observation window along temporal frames and conduct classification within each window using multiple features.  
Inspired by recent works on object detection from still images~\cite{ren2015faster,girshick2014rich,yang2021pdnet,yang2022vgvl}, the idea of generating object proposals has been borrowed to perform action detection from video sequences~\cite{siva2011weakly,escorcia2016daps,jain2014action,gkioxari2015finding,chen2014actionness,heilbron2016fast,shou2016action,peng2016multi,singh2017online}.
Some of these methods~\cite{jain2014action,gkioxari2015finding,chen2014actionness,zhao2017temporal} produce spatio-temporal object volumes to perform spatio-temporal detection of simple or cyclic actions. 
Different from these methods, we propose an end-to-end self-attention model for \tdtal.

\noindent
{\bf Transformers.} 
Transformer architectures \cite{vaswani2017attention} have recently demonstrated tremendous performance in sequence modeling tasks such as machine translation \cite{vaswani2017attention,gao2021scalable}, language modeling \cite{Devlin2019BERTPO,Brown2020LanguageMA}, video captioning \cite{zhou2018end,sun2019videobert}, etc. 
Among the different variants of transformer architectures, we find most relevant, the recent line of work that predicts a set output, based on modeling the context of the input. 
Recently, DETR~\cite{carion2020end} formulated object detection as an end-to-end task. 
Follow-up work, Deformable-DETR~\cite{zhu2020deformable} improves the efficiency and effectiveness of this architecture by employing sparse attention. 
In a similar vein, Activity Graph Transformers \cite{nawhal2021activity} employ sparse graph attention for a 2D-TAL task, and report performance improvements. 
A previous (non-transformer) method for 2D-TAL \cite{xu2020g} also modeled context as a graph, and reported improvements.  
Although our task is different from these works, the architecture, training methods, and insights about the use of sparse attention for large contexts, are all relevant. 
\section{New benchmark}
\label{sec:benchmark}

\noindent
\textbf{Motivation.} 
\label{sec:benchmark_motivation} 
State-of-the-art \tdtal approaches tackling \tdtal typically train their approaches on PKU-MMD \cite{liu2017pku}, and achieve excellent performance ($> 80.0\%$ mAP). 
In a proof-of-concept experiment, we train methods on fractions of the large training set, which we expected to be a more challenging setting. 
Surprisingly, \algoName~achieves close to state-of-the-art performance when trained on just half of the training set. 
This suggests that the motions in PKU-MMD are largely similar. 
We argue that to drive meaningful progress in the 3D-TAL task, a more realistic, and challenging benchmark is necessary. 

\begin{table*}[t!]
  \centering
    \caption{
    Comparison of \algoName with previous methods on the PKU-MMD dataset (mAP@tIoU$= 0.5$). In parentheses are the fractions of the training set that are used.
    }
  \begin{tabular}{p{4.2cm}<{\centering}|p{3.3cm}<{\centering}|p{3.3cm}<{\centering}}
    \toprule
  {\bf Method} & Cross-view & Cross-subject	\\
    \midrule
    \texttt{Beyond-Joints}~\cite{beyond_joints} ($100\%$) & $91.1$ & $81.1$ \\
    \texttt{Cui et al.}~\cite{cui2020skeleton} ($100\%$) & $93.3$ & $83.5$ \\
    {\bf \algoName} ($100\%$) & $\bm{94.6}$ & $\bm{93.2}$  \\
    \toprule
    \texttt{Beyond-Joints}~\cite{beyond_joints} ($50\%$) & $89.6$ & $64.4$ \\
    \algoName ($50\%$) & $94.3$ & $93.0$  \\
    \toprule 
    \texttt{Beyond-Joints}~\cite{beyond_joints} ($10\%$) & $80.1$ & $35.0$ \\
    \algoName ($10\%$) & $91.9$ &  $64.1$ \\
    \bottomrule
  \end{tabular}
  \label{tab:pku_mmd}
\end{table*}

\noindent
\textbf{BABEL-TAL-60 (BT60).} 
\label{sec:dataset_introduction}
We propose the recently released BABEL dataset \cite{BABEL:CVPR:2021} as a new \tdtal~benchmark. 
BABEL is diverse, containing more than $250$ actions whose frequency follows a long-tailed distribution. 
We first consider a 60-class subset of BABEL as BT60, where many classes have $<5$ training sequences. 
Unsurprisingly, this is extremely challenging, and \algoName achieves mAP$@0.5 = 0.01\%$. 
Learning to localize actions in this few-shot setting is an interesting open research problem for future research. In the rest of the paper, we employ the below experimental setting which is of moderate difficulty.

\noindent
\textbf{\datasetName (BT20).} 
To construct a benchmark of moderate difficulty, we identify the most common actions in BABEL and categorize them into 20 classes. 
Choosing the most frequent actions ensures sufficient training data for algorithms. 
We find that among the most frequent actions, there are a few fine-grained categories. 
For instance, semantically similar actions like `run' and `jog' belong to different categories in BABEL. 
Due to the challenging nature of the \tdtal task which requires \emph{both} localization and recognition, we consider slightly more coarse-grained categories to ease the learning problem. 
We create a new set of $20$ actions, where each action in BT20 is a union of semantically similar BABEL action categories. 
Each class has varying complexity, intra-class variance, and amount of data. 
This subset, called BT20, contains $5727$ mocap sequences with a duration of $\sim 17$ hours (see~\cref{tab:comp_dataset}). 

Similar to BABEL, BT20 is a benchmark with a long-tailed distribution of actions. In the Sup. Mat., we provide stats regarding the action categories in BT20. 
The training and validation splits for BT20 are identical to BABEL. 

\section{Method}
\label{sec:method}

\noindent
\label{sec:formulation}
3D Temporal Action Localization (\tdtal) involves predicting the action label, start time and end time of every action occurring in a 3D motion sequence. 
Specifically, given a 3D motion sequence $\mathbf{x}_i$, the goal is to predict the set of $N_i$ action spans 
$\Psi_i =\{\psi^{n} = (c^{n}, t^{n}_s , t^{n}_e)\}^{N_i}_{n=1}$, 
where $c^{n}$ is the action category of the $n$-th span and $(t^{n}_s, t^{n}_e)$ denote the start, end times of the $n$-th span. 

\subsection{Input representation}
\label{sec:inp_rep}
\noindent
\textbf{Frame representation.} 
The positions of the joints of a person (in 3D Cartesian coordinates), represents the pose of a person over time. 
\algoName considers 22 joints (all except hands) in the SMPL body model \cite{SMPL-X:2019} skeleton format. 
Thus, each frame in the 3D motion sequence is represented by a pose feature $\theta_t \in \mathbb{R}^{22 \times 3}$. 
For efficient learning, we reduce the variance in the input data. 
Following Shi et al.~\cite{shi2019two}, we rotate the skeleton such that the hip bone is parallel to the x-axis, and the shoulder bone to the y-axis. We subtract the position of the pelvis from all other joint positions.

Prior works in transformer-based video-captioning \cite{sun2019videobert} and 2D-TAL \cite{nawhal2021activity} provide features from an action recognition model as input to the model. 
Although this increases the computation and memory requirements, we explore these for \tdtal with limited gains (see Sup. Mat.). 

\noindent
\textbf{Sequence representation.} 
Transformers accept only fixed-length inputs but motion sequences vary in length. 
The standard approach is to implement large models that can accommodate most sequences in the dataset. 
This requires large memory and computation. Further, a large temporal dimension causes the initial attention to be diffuse and ineffective \cite{zhu2020deformable}, 
which further delays convergence. 

A solution is to fix the input size of the model, and down-sample a long (or up-sample a short) sequence of poses. 
However, this input representation loses information about the velocity of movement which may be detrimental to accurate recognition. 
We handle this by utilizing a short snippet of $N_f$ contiguous poses as input to each element (time-step). 
Each element has access to local temporal information, and the temporal size of the model is reasonable. 
The overall 3D joint position features $\bm{x}_\theta \in \mathbb{R}^{22*3*N_{f} \times T}$. 

While down-sampling long sequences, the frames within each snippet are retained and frames between contiguous sequences are dropped. 
The input representation for short sequences consists of neighboring snippets having redundant frames, i.e., snippets overlap. 

\begin{figure*}[t!]
  \centering
    \includegraphics[width=\linewidth]{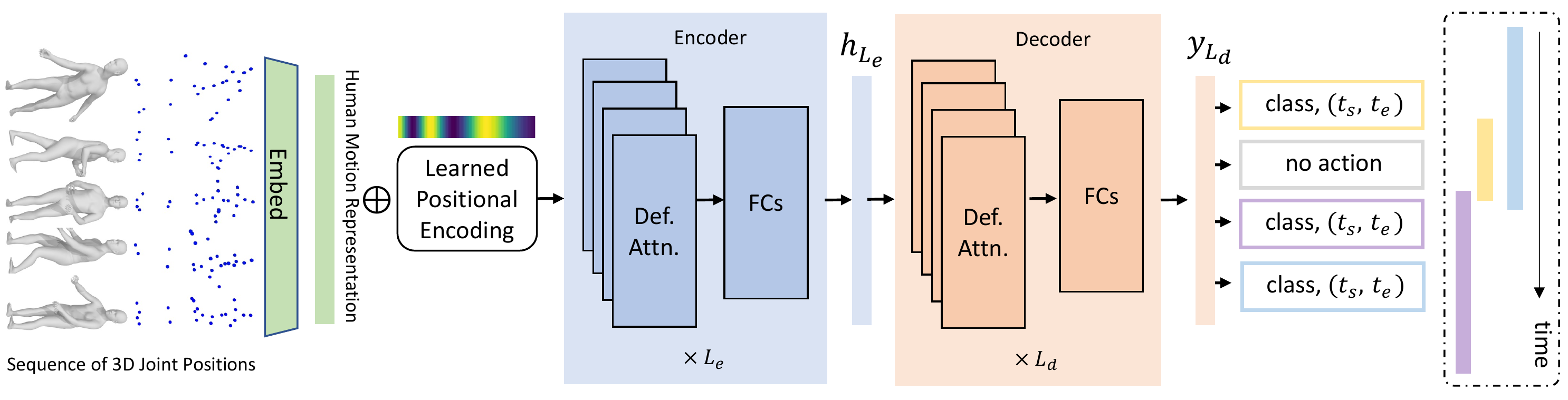}
  \caption{
    Given a sequence of 3D poses, \algoName outputs a set of action spans. 
    A projection of the 3D pose features is summed with positional (time) information, and input to the transformer. 
    The transformer encoder models the global context across all positions to produce a feature $\bm{h}_{L_e}$ that encodes action span information in the motion. 
    The decoder transforms a fixed set of span queries into corresponding span predictions by modeling the span information in the sequence $\bm{h}_{L_e}$. 
    The decoder models context across different spans to produce a set of span output features $\bm{y}_{L_d}$. 
    Given the output feature for each span, the prediction head categorizes the action (\texttt{class}), and regresses to its start ($t_s$) and end time ($t_e)$. 
    \algoName employs a sparse attention mechanism (\texttt{Def.Attn.}) \cite{zhu2020deformable} to model context. 
    }
  \label{fig:framework}
\end{figure*}

\subsection{\algoName Architecture}
\label{sec:arch}

Figure \ref{fig:framework} shows the overall architecture of our approach. 
In a single forward pass, \algoName identifies and localizes all actions in a 3D motion sequence. 
The output is a set of spans ($c$, $t_s$, $t_e$), which enables \algoName to model simultaneous actions. 
Further, \algoName is trained end-to-end. 
We describe each component in detail:

\noindent 
\textbf{Transformer Encoder.}
\label{sec:enc}
The sequence of 3D joint position features $\bm{x}_\theta$ are each embedded to a feature space of size $C$, resulting in $\bm{x} \in \mathbb{R}^{C \times T}$. 
Each position $t$ is associated with a positional encoding $\bm{pe}_t \in \mathbb{R}^C$, where $t \in \{1, \dots, T\}$. 
The sum of $\bm{pe}$ with features, $[\bm{x}_t + \bm{pe}_t]_{t=1}^{T}$ are input to $E_1$. 

A single encoder layer $E_\ell$ involves the following operations: (1) Self-attention computation, (2) Concatenation of features from different heads, (3) Projection back into the original feature dimension (4) Residual connection \cite{He2016DeepRL} (5) Layer Normalization \cite{ba2016layer}.  
The self-attention operation effectively models context across features in all $T$ positions. 
Each transformer head projects the features from the previous position into a different feature space, capturing different signals. 
In an aggregation step, features from all heads are concatenated and projected back into the original feature dimension, followed by residual connection.
Overall, the encoder feature $h_{L_e} = E_{L_e} \circ \dots \circ E_1({\cdot})$ is the resulting output from all the layers. 

\noindent \textbf{Transformer Decoder.} 
\label{sec:dec}
The decoder $\mathbf{D}$ contains $H$ heads, and $L_d$ layers. 
Following previous work \cite{carion2020end,zhu2020deformable,nawhal2021activity}, we provide action query embeddings as input to the decoder. 
Action query embeddings are learned positional encodings. The number of embeddings $N_a$ is much larger than the maximum number of action spans for a motion in the dataset. 

The decoder learns to transform the input action query embeddings into action span predictions for the motion sequence. 
Each decoder layer $D_\ell$ performs two attention computations: self-attention, and cross-attention. 
Similar to the encoder, self-attention involves modeling context across different positions of the decoder. 
Cross-attention involves attending to the encoder feature $h_{L_e}$ (key), using the decoder features as queries. 
Overall, the decoder outputs an action query representation $y_{L_d} = D_{L_d} \circ \dots \circ D_1({\cdot})$. 

We further implement two decoder variants -- iterative bounding box refinement and two-stage decoder \cite{zhu2020deformable}, resulting in performance gains (see Sup. Mat.). 

\noindent \textbf{Deformable Attention.} 
\label{sec:def_att} 
In transformers, the representation of an element at a successive layer is a function of the element itself, and its context. 
In `regular' self-attention \cite{vaswani2017attention}, \emph{all} elements at a level are considered the context. 
Nahwal et al.~\cite{nawhal2021activity} observe that GAT\cite{velickovic2018graph} is more effective in the 2D-TAL task. It defines the neighbors of an element based on the learned attention weights between the element and others. 
With deformable attention \cite{zhu2020deformable}, the model predicts a small, fixed set of relative positions of the context that is relevant for an element. This is more flexible, as the neighbors are not constrained to be $K$ unique positions. 

To describe deformable attention (DA), we borrow notations from Zhu et al.~\cite{zhu2020deformable}. 
Consider a feature $\bm{x} \in \mathbb{R}^{C\times T}$ at a certain transformer layer. 
Given the feature at a particular position $\bm{z}_q \in \mathbb{R}^{C}$, regular attention mechanisms involve attending to all positions in $\bm{x}$. 
However, DA aims to attend to a small set of relevant elements from $\bm{x}$. 
Concretely, given a reference location $p_q$ (the $q$-th position of the query feature $\bm{z}_q$), 
DA identifies $K$ relevant sampling locations from $\bm{x}$. 
Since sparsity is the goal, $K \ll T$. 
The sampling locations are predicted relative to the reference location $p_q$, and hence denoted $\Delta p_{qk}$. $k$ indexes the sampled keys. 
The model learns parameters $\bm{W}^V$ to project the features of the sampled keys $\bm{x}(p_q+\Delta p_{qk})$. 
DA for one attention head is expressed in \cref{equ:deformable}. 
\begin{equation}
\label{equ:deformable}
\textrm{DA}(\bm{z}_q, p_q, \bm{x}) = \sum_{k=1}^K  A_{qk} \cdot \bm{W}^{V} \bm{x}(p_q+\Delta p_{qk})
\end{equation}
Here, $A_{qk}$ denotes the scalar attention weight of the $k$-th sampled key. 
$A_{qk}$ is in the range $[0, 1]$, normalized over $K$ to sum to $1$. 
Both $A_{qk}$ and $\Delta p_{qk}$ are obtained by a linear projection over $\bm{z}_q$. 
$\Delta p_{qk}$ is a real valued number. A bilinear interpolation is performed to compute $\bm{x}(p_q+\Delta p_{qk})$. 

The overall feature with $M$ attention heads is a weighted-sum over the DA representation presented in \cref{equ:deformable}. 
Please refer to Zhu et al.~\cite{zhu2020deformable} for more details. 

\noindent \textbf{Prediction Heads.} 
\label{sec:pred_head}
Our prediction heads are similar to Carion et al.~\cite{carion2020end}. 
A regression network, and a recognition network operating on each of the $N_a$ action queries from $y_{L_d}$, comprise the prediction heads. 
The action query representation $y_{L_d} \in \mathbb{R}^{C \times N_a}$ from the decoder contains information regarding the span of actions in the motion. 
The regression network consists of a 3-layer fully connected network (FC) with ReLU activation that predicts the start time $\hat{t}_s$ and end time $\hat{t}_e$ of the action. 
The recognition network is a single fully connected with a softmax function that scores the set of actions (which includes a `no action' class). 
Note that the number of ground truth action spans $|\Psi_i|$ is variable, and smaller than $N_a$, the number of predictions.
The model learns to predict the `no action' class when the desired number of action spans have been predicted. 

\subsection{Loss functions}
\label{sec:loss} 
The key idea with \algoName is to learn to jointly perform localization and recognition. 
Thus, \algoName is trained with two objectives -- for classification and regression. 
We handle the differing number of spans between the ground-truth and predictions by computing a bipartite matching between \cite{carion2020end,Stewart2016EndtoEndPD}. 

\noindent
\textbf{Bi-partite Matching.} 
We ensure that the two sets have the same cardinality by augmenting the ground-truth $\Psi_i$, with `no action' spans, such that the augmented ground-truth $\Tilde{\Psi}_i$ has $N_a$ spans. 
We compute an optimal match between the augmented ground-truth $\Tilde{\Psi}_i$ and prediction $\hat{\Psi}_i$. 
We search for the optimal permutation $\sigma^*$ among the set of all permutations $\Sigma_{N_a}$, that has the lowest matching cost. 
For clarity, going forward, we drop the data sample subscript $i$. 
\begin{equation}
\sigma^* = \textrm{argmin}_{\sigma \in \Sigma_{N_a}} \sum_{n=1}^{N_a} \mathcal{L}_\textrm{match}(\Tilde{\Psi}^{n}, \hat{\Psi}^{\sigma(n)})
\end{equation}
This cost is computed efficiently via the Hungarian algorithm. $\mathcal{L}_\textrm{match}$ depends on both the recognition, and regression losses. 

\noindent
{\bf Regression loss.} 
$\mathcal{L}_\textrm{span}$ measures the localization similarity between the predicted and ground-truth action spans. 
$\mathcal{L}_\textrm{span}$ has two terms -- $\mathcal{L}_1$ loss which accounts for the span duration, and $\mathcal{L}_{iou}$ which is invariant to the span durations. 
$\mathcal{L}_\textrm{span}$ is a weighted combination between these terms, weighted by scalar hyperparameters $\lambda_{iou}$ and $\lambda_{L_1}$. 
Considering that we compute $\mathcal{L}_\textrm{span}$ for all match permutations $\sigma$:
\begin{equation}
    \mathcal{L}_\textrm{span} = \lambda_{iou}\mathcal{L}_{iou}(s^{n}, \hat{s}^{\sigma(n)}) + \lambda_{L_1}||s^{n} - \hat{s}^{\sigma(n)}||_1,
\label{equ:local}
\end{equation}
where $s^n$ is the start and end times $[t_s^n, t_e^n]$ of the $n$-th span. 

\noindent
{\bf Classification loss.} 
In an attempt to effectively model the heavy class imbalance, we experiment with different classification losses -- cross-entropy loss, focal loss $\mathcal{L}_{F}$~\cite{lin2017focal}, and class-balanced focal loss $\mathcal{L}_{CB}$~\cite{cui2019class}. 
Focal loss up-weights the cross-entropy loss for inaccurate predictions, resulting in a larger training signal for difficult samples. 
Class-balanced focal loss incorporates a class-weighting term, called `effective number' of samples. This is a non-linear function of the class frequency, which accounts for the uniqueness of training samples. 
\algoName utilizes $\mathcal{L}_{CB}$ which performed best in our experiments (see Sup. Mat.). 

Given predicted class scores $\bm{z}$, we define $\Tilde{z}^j$ as
\begin{equation}
\Tilde{z}^j = \begin{cases} z^j, & \text{if}\ j = c^n \\
-z^j,  & \text{otherwise}
\end{cases}
\end{equation}
Further, we define $\hat{p}^j= \textrm{sigmoid}(\Tilde{z}^j)$. 
We compute the  $\mathcal{L}_{CB}$ between the ground truth-class $c^n$ and the predicted class scores under match permutation $\hat{p}_{\sigma(n)}(c^n)$. \\
Concretely, 
$\mathcal{L}_{CB}(c^{n}, \hat{p}_{\sigma(n)}(c^{n})) = $
\begin{equation}
    -\frac{1-\beta}{1-\beta^{\mathrm{count(c^n)}}}\sum_{j=1}^C\left(1-\hat{p}^j_{\sigma(n)}(c^{n})\right)^\gamma\log\left(\hat{p}^j_{\sigma(n)}(c^{n})\right)
    \label{equ:class}
\end{equation} 
where $\mathrm{count(c^n)}$ is the frequency of the ground-truth class $c^n$, $\beta \in [0, 1)$ and $\gamma$ are scalar hyperparameters, and $C$ is the total number of classes. 

\noindent
{\bf Overall objective.} 
The bipartite matching loss $\mathcal{L}_\textrm{match}$ is a sum of the classification and regression losses:
\begin{equation}
    \begin{aligned}
\mathcal{L}_\textrm{match}(\Tilde{\Psi}^{n}, \hat{\Psi}^{\sigma(n)}) = \mathcal{L}_{CB}(c^{n}, \hat{p}_{\sigma(n)}(c^{n})) + \mathcal{L}_\textrm{span}(s^{n}, \hat{s}^{\sigma(n)}) \\ 
    \end{aligned}
\end{equation}

After obtaining the optimal permutation $\sigma^*$ based on the lowest $\mathcal{L}_\textrm{match}$, we compute the Hungarian loss $\mathcal{L}_H$ for this optimal matching. 
$\mathcal{L}_H$ over all the matched pairs of spans is defined as: 
\begin{equation}
    \mathcal{L}_H = \sum_{n=1}^{N_a} \mathcal{L}_\textrm{match}(\psi^{n}, \hat{\psi}^{\sigma^*(n)}) 
\end{equation}
\algoName is trained end-to-end, and jointly for both recognition and localization with $\mathcal{L}_H$ as its objective.

\section{Experiments}
\label{sec:exp}

\subsection{Comparison with other methods} 

\noindent{\bf \texttt{Beyond-Joints.}} 
On the current \tdtal benchmark PKU-MMD \cite{liu2017pku}, Wang and Wang \cite{beyond_joints} achieve close to state-of-the-art performance, with mAP @ $0.5$ tIoU $=81.1$ (cross-subject)\footnote{
The cascaded approach involving RNN-based localization, followed by RNN-based recognition, by Cui et al.~\cite{cui2020skeleton} achieves the state-of-the-art with mAP $=83.5$. 
}. 
\texttt{Beyond-Joints} consists of an RNN-based per-frame action-classifier, whose predictions are aggregated by a localization algorithm involving multi-scale sliding windows. 
We benchmark the performance of their publicly available implementation\footnote{\url{https://github.com/hongsong-wang/Beyond-Joints}} on BT20. 

\noindent{\bf \texttt{3D-ASFD.}}~\cite{lin2021learning} 
ASFD is a purely anchor-free 2D temporal localization method with a saliency-based refinement module to gather more valuable boundary features for each proposal with a novel boundary pooling. We re-implement ASFD and adapt it to 3D input.

\noindent{\bf \texttt{3D-TSP.}}~\cite{alwassel2021tsp}
Alwassel et al. proposed a 2D-TAL paradigm for clip features that not only trains to classify activities but also considers global information to improve temporal sensitivity. We modify the author’s implementation to use the same 3D features as our model.

\noindent{\bf \texttt{3D-GTAD.}} 
G-TAD \cite{xu2020g} is a state-of-the-art 2D-TAL approach whose key idea is to effectively model the context around a short span of video. 
A video is considered a graph, with short time-spans as nodes. 
Temporally adjacent spans are joined by edges. Edges also encode the relationship between spans and the background. 
We adapt this 2D approach to utilize 3D features as input, and benchmark its performance on BT20. 

\noindent{\bf \texttt{\algoName w/ GAT.}} 
Similar to G-TAD, Activity Graph Transformer \cite{nawhal2021activity} is a recent 2D-TAL approach that aims to model non-sequential temporal correlations by considering a video as a graph. 
It employs a Graph Attention (GAT) mechanism \cite{velickovic2018graph}. 
The graph-attention models the neighborhood of a particular node, where neighbors are nodes with the largest learned edge weights. 
Instead of the regular self and encoder-decoder attention in the transformer \cite{vaswani2017attention}, we employ the graph self-attention and graph-graph attention \cite{nawhal2021activity} in \algoName, for \tdtal. 

\noindent{\bf\algoName \texttt{base}} is the base architecture of \algoName without additional improvements such as iterative bounding box refinement, and a two-stage decoder. The full algorithm, {\bf \algoName}, includes these `bells and whistles'.

\begin{table*}[t!]
  \centering
  \caption{
    Performance of a prior method, strong baselines, and our approach \algoName on the BT20 benchmark. 
    Each column indicates a tIoU threshold in the range $[0.1, 0.9]$. 
    A cell presents the mAP at that tIoU threshold. The final column presents Avg.~mAP over all thresholds. 
    \algoName outperforms all other methods at almost all tIoU thresholds. Gains are larger for lower tIoU thresholds, implying better recall.
    }
  \begin{tabular}{lcccccccccc}
    \toprule
  {\bf Method / tIoU} & $\bm{0.1}$	& $\bm{0.2}$	& $\bm{0.3}$	& $\bm{0.4}$	& $\bm{0.5}$	& $\bm{0.6}$	& $\bm{0.7}$	& $\bm{0.8}$	& $\bm{0.9}$   & {\bf Avg.}	\\
    \midrule
    \texttt{Beyond-Joints} \cite{beyond_joints} & $14.3$ & $13.6$ & $13.3$ & $12.3$ & $11.4$ & $10.5$ & $8.92$ & $6.24$ & $4.13$ & $10.5$ \\
    \texttt{3D-ASFD}~\cite{lin2021learning} 
    & $24.2$ & $23.1$ & $22.6$ & $22.2$ & $21.9$ & $20.4$ & $18.9$ &  $12.2$ & $9.01$ & $19.3$ \\
    \texttt{3D-TSP}~\cite{alwassel2021tsp} 
    & $26.9$ & $25.6$ & $24.1$ & $23.0$ & $22.5$ & $20.4$ & $17.1$ & $13.0$ & $10.1$ & $20.3$ \\
    \texttt{3D G-TAD}~\cite{xu2020g} & $25.1$ & $24.1$ & $23.9$ & $23.0$ & $22.1$ & $21.1$ & $18.5$ & $14.1$ & $11.8$ & $20.4$ \\
    \algoName \texttt{w/ GAT} & $27.3$ & $26.0$  & $25.7$  & $24.5$ & $23.4$ & $21.5$ & $19.4$ & $15.9$ & $\bm{12.4}$ & $21.9$  \\
    \algoName \texttt{base} & $43.5$ & $41.1$ & $41.0$ & $38.2$ & $35.1$ & $\bm{30.5}$ & $\bm{23.7}$ & $\bm{16.4}$ & $9.99$ & $31.1$ \\
    {\bf \algoName} & $\bm{46.6}$  & $\bm{45.5}$  & $\bm{43.0}$  & $\bm{40.2}$  & $\bm{36.0}$  & $\bm{30.5}$  & $\bm{23.7}$  & $15.9$  & $9.78$ & $\bm{32.0}$ \\
    \bottomrule
  \end{tabular}
  \label{tab:comp}
\end{table*}

\subsection{Results}

\noindent \textbf{Metric.} 
We train models on the train set and report results after evaluating on the validation set of \datasetName and report the mean Average Precision (mAP) at various temporal IoU thresholds. 
Prior TAL benchmarks vary in the specific tIoU thresholds that mAP is computed at -- ActivityNet-1.3 \cite{xiong2016cuhk} @ tIoU$=\{0.5, 0.75, 0.95\}$, THUMOS-14 \cite{THUMOS14} @ tIoU$=\{0.3, 0.4, 0.5, 0.6, 0.7\}$, and PKU-MMD \cite{liu2017pku} @ tIoU$= 0.5$. 
To be comprehensive, we report mAP over 9 different tIoU thresholds $[0.1: 0.9: 0.1]$ on the \datasetName benchmark. 
In ablation studies and analyses, for simplicity, we report mAP @ tIoU $=0.5$. 
Please see Sup. Mat. for details regarding training, architecture, and hyper-parameters. 

\noindent \textbf{Does \algoName outperform prior approaches?}
We observe that the \texttt{3D-GTAD} baseline outperforms the RNN-based prior work \texttt{Beyond-Joints} \cite{beyond_joints}. 
This suggests that modeling global context, including background, is more effective than only modeling local temporal neighborhoods. 
Further, all the transformer encoder-decoder-based approaches, which can also model long-term context, outperform \texttt{Beyond-Joints} by larger margins. 
Specifically, our approach \algoName outperforms the prior method \texttt{Beyond-Joints} by $13.9$ points (mAP @ tIoU$=0.5$). 

\noindent \textbf{Does the attention mechanism matter?} 
The attention mechanism makes a crucial difference in the overall performance of \algoName. 
While both GAT and Deformable Attention (DA) are sparse attention mechanisms, the flexibility of DA improves the performance of \algoName significantly. 

\noindent \textbf{Do bells and whistles matter?} 
We observe that Iterative bounding box refinement and two-stage decoders \cite{zhu2020deformable} further improve the performance of the algorithm. 
The two-stage decoder first proposes action spans, which are then provided as action query features to the decoder. 
These improvements present an interesting trade-off between the simplicity of the model, and performance. 
It also suggests a potential direction for improvement -- action query features. 

\noindent \textbf{Large gains from \algoName at lower tIoU.} 
Overall, \algoName outperforms the competition across all tIoU thresholds. 
However, the gains are comparatively large for lower tIoU thresholds. 
This indicates that \algoName has a relatively better recall. This of course, comes at the cost of precision. In downstream applications, we observe that applying Non-Maximal Suppression (NMS) on the predictions improves results qualitatively (despite the drop in mAP). 

\noindent \textbf{\texttt{3D-GTAD} vs. \algoName \texttt{w/ GAT.}} 
It is interesting to note that these methods which have a similar basic graph formulation, perform similarly. 
While there are many points of difference between the two models that make it difficult to precisely identify the sources of difference, there are two main differences: 
(1) \texttt{3D-GTAD} representations are learned via graph convolutions that model the neighborhood context, while \algoName \texttt{w/ GAT} employs graph attention. Prior work suggests that GAT outperforms GCNs in certain tasks \cite{velickovic2018graph}, so it is possible that this is a source of improvement. 
(2) Given a (feature) graph, \texttt{3D-GTAD} uses pre-defined anchors to perform localization. In contrast, \algoName \texttt{w/ GAT} learns to predict the start and end of spans in end-to-end manner. 
Fine-grained control over spans appears to improve performance on \tdtal. This is further validated by the relatively small improvement due to the bells and whistles discussed above. 

\subsection{Human Experiments}

\noindent \textbf{Motivation -- limitation of mAP metric.} 
Labeling an action and marking its precise start and end in the motion, is a highly subjective task. 
It involves grounding one's semantic concept of `action' and its precise duration in the physical movement. 
Invariably, there exists disagreement between human labels for the same motion. 
However, the definition of `ground-truth' labels typically only considers an annotation from one person. 
Thus, it is possible that a prediction that is accurate with respect to labels from one annotator, is penalized when a different annotator's labels are considered ground-truth. 
This implies that although the upper-bound of the mAP metric is $1.0$, an approach might perform quite well even if its mAP $\ll 1.0$. 

\noindent \textbf{Human performance (mAP).} 
To obtain a concrete reference for `good' mAP performance, we compute the `human mAP' for BABEL sequences containing different (unique) human annotations. 
There are $67$ motion sequences that have multiple annotations, with $256$ action spans overall, containing one of the  $20$ actions in \datasetName. 
For each motion, we consider one annotation as the ground-truth, and the other human annotations as predictions. 
Interestingly, on these sequences, the `human mAP' @ tIoU$=0.5$ is only $0.27$. 
While this analysis is only on a small subset of data, we do observe the significant variance in human annotations qualitatively (see visualizations in Sup. Mat.). 

In the image captioning task, variance in semantic labels is handled via metrics that consider labels from multiple humans \cite{vedantam2015cider}. 
Since this solution involves significant labeling overhead, 
we directly compare the performance of \algoName vs. \texttt{Beyond-Joints} via a head-to-head human evaluation. 

\noindent \textbf{Experimental setting for human evaluation.} 
We simultaneously present an evaluator with two videos of the same motion\footnote{Human evaluators are recruited from the crowdsourcing platform Amazon Mechanical Turk: \url{https://www.mturk.com/}. We provide the task interface in the Sup. Mat.}. 
We overlay the two videos with labels from \algoName, \texttt{Beyond-Joints}, and \texttt{3D-GTAD} (in random order across trials).  
The evaluator answers the following question -- `Which labels better match the motion (left or right)?' 
We collect votes from $5$ unique evaluators for each motion sequence to account for subjectivity. 
We compare labels from \algoName head-to-head with labels from \texttt{Beyond-Joints}. 
We observe that human evaluators prefer labels from \algoName $69.20$\% of the time; significantly more than \texttt{Beyond-Joints} ($30.80$\%). We also observe that \algoName outperforms \texttt{3D-GTAD} -- $55.31\%$ vs. $44.69\%$. 
It is encouraging that the mAP metric and human evaluation appear to be correlated. The differences between \texttt{Beyond-Joints} and \algoName are larger than differences between \texttt{3D-GTAD} and \algoName under both metrics. 

\subsection{Analysis}
\algoName~achieves highest  Average Precision (AP) for the actions `run' and `walk' (see Fig. in Sup. Mat.).
While performance on `run' drops gradually with increasing tIoU, the drop is more pronounced for `walk'. This indicates that the localization is good for `walk', and extremely good for `run'. I.e., the overlap between predicted and ground-truth spans is large for `run'. 
Interestingly, we find that \algoName achieves reasonable AP for `place something', despite not having access to hand or finger articulation data. 
The Sup. Mat. contains further performance analysis, error-analysis with a confusion matrix, ablation studies, and visualization of results from different approaches. 

\section{Conclusion}
\label{sec:conclusion}

3D Temporal Action Localization (\tdtal) is an important computer vision task with many applications. 
We present \algoName, a novel \tdtal approach that learns to jointly perform localization and recognition. 
Given a 3D sequence of joint positions, \algoName predicts action spans in a single forward pass. 
\algoName outperforms all baselines in automatic metrics and human evaluation. 
We observe the saturating performance of methods on the current \tdtal benchmark, and propose BT20, a new, more challenging benchmark for \tdtal. 
\algoName outperforms existing methods on BT20 as well, but the overall performance indicates that there is significant room for improvement for \tdtal methods on the benchmark. 
The availability of body-surface meshes in BT20 -- unique among \tdtal datasets -- could be a useful signal for future methods that reason about human-scene/human-object interactions. 
We believe that the method, code, and findings in this work will be useful to the community. 

\clearpage
\section*{Appendices}
\addcontentsline{toc}{section}{Appendices}
\renewcommand{\thesubsection}{\Alph{subsection}}
\renewcommand{\thesection}{\Alph{section}}
\setcounter{section}{0}
In this document, we present additional details, quantitative results, and qualitative analysis of model predictions. 
The contents are as follows: 
\begin{itemize}
    \item \cref{sec:dataset}: Dataset stats.
    \item \cref{sec:imp_details}: Implementation details.
    \item \cref{sec:sources_error}: Analysis -- Sources of error.
    \item \cref{sec:analysis}: Per-class performance.
    \item \cref{sec:inp_feat}: Experiments on Input features.
\end{itemize}

\section{Dataset}
\label{sec:dataset}
The BABEL dataset \cite{BABEL:CVPR:2021} consists of more than $250$ actions whose frequency follows a long-tailed distribution. 
To ensure that the learning methods have access to sufficient data, the BABEL action recognition benchmark only considers a subset of actions for the task. 
Similarly, for the 3D Temporal Action Localization (\tdtal) task, we only consider the most frequent actions in BABEL. 

We create the BABEL-TAL-60 (BT60) benchmark for \tdtal the 60-class action recognition subset of BABEL presented by Punnakkal et al.~\cite{BABEL:CVPR:2021}. 
The action recognition task involves predicting the action class of the given `trimmed' span of movement. The movement spans in the BABEL-60 subset belong to the most frequent 60 actions in BABEL. 
For the BT60 \tdtal benchmark, we consider all the original motion \emph{sequences} from which spans are utilized for BABEL-60. 
Given the full mocap sequence, the task is to localize and recognize all of the 60 actions. 
This is extremely challenging (see Sec.~3 in the main paper), especially due to many infrequent actions. 
Due to the expensive nature of (dense) annotation required for TAL tasks, we believe that learning efficiently from limited amounts of fully annotated data, is an important problem for the community to solve in the long run. 

\begin{table*}[t]
  \centering
  \caption{
    \datasetName (\btt) has $20$ actions (column 1), where each is a superset of BABEL actions (column 2). The number of samples in the training and validation set of BT20, are in columns 3 and 4. 
  }  
  \begin{tabular}{l|c|c|c}
\toprule
{\bf \btt} & {\bf BABEL} & {\bf \# Train}  & {\bf \# Val}	\\
\midrule
walk & walk & 4671 & 1591 \\
stand & stand & 4193 & 1615 \\
turn & turn, spin & 2044 & 837 \\
step & step & 1097 & 458 \\
run & jump, hop, leap & 750 & 261 \\
lift something & take/pick something up, lift something & 709 & 244 \\
jump & jump & 700 & 214 \\stretch & stretch, yoga, exercise/training & 601 & 237 \\
sit & sit & 512 & 181 \\
place something & place something & 510 & 161 \\ 
bend & bend & 468 & 161 \\
throw & throw & 460 & 128 \\
kick & kick & 347 & 144 \\
touch body part & scratch, touching face, touching body parts & 310 & 137 \\
grasp object & grasp object & 247 & 105 \\
hit & hit, punch & 206 & 105 \\
catch & catch & 193 & 59 \\
dance & dance & 189 & 89 \\
greet & greet & 179 & 70 \\
kneel &  kneel & 102 & 25 \\
    \bottomrule
  \end{tabular}
  \label{tab:dataset_intro}
\end{table*}

To encourage rapid progress in the \tdtal task in the short run, we construct a moderately challenging benchmark called BT20. 
BT20 is derived from the most frequent actions in BABEL-60 by grouping related fine-grained actions to form a coarser set of action classes. 
In \cref{tab:dataset_intro}, we list each action in \btt, the corresponding BABEL categories that constitute the action, and the number of samples in the training and validation set of \btt. 
Note that the frequency of coarse actions in \btt also follow a long-tailed distribution. 

The training and validation splits for \btt are identical to the one provided by BABEL\footnote{\url{https://github.com/abhinanda-punnakkal/BABEL}}. The splits are randomly generated based on the sequence. 

The PKU-MMD \cite{liu2017pku} benchmark considers the cross-view and cross-subject evaluation settings. 
However, this is in applicable for BT60 and BT20, which only have a single evaluation setting each, on their respective validation splits. 

\noindent
{\bf Cross-view.} 
A 3D skeleton (joint positions) in a given view can be rotated with respect to a reference co-ordinate system to generate a canonical view. Indeed, we normalize sequences such that the spine is parallel to the z-axis, and hip bone is parallel to the x-axis (described in Sec.~4.2 in the main paper). 
This pre-processing renders the cross-view setting inconsequential as a generalization measure. 

\noindent
{\bf Cross-subject.} 
The mocap sequences from the AMASS dataset \cite{AMASS:ICCV:2019} are obtained from a large number of subjects ( $> 346$). This reduces the risk of a learning method over-fitting to a few subjects. 
Specifically, the joint position features are computed from mocap sequences using the SMPL body model \cite{SMPL:2015}. 
We assume a subject of `mean body shape' while computing the skeleton joint positions. 
This is a useful approximation -- the consistency in body shape across sequences ensures that there is no variation in the skeleton across samples. 
While this marginally reduces the fidelity of the joint position features (the exact position of a joint depends on the body shape), it also prevents the model from having to learn to generalize to different body shapes. 

\section{Implementation details}
\label{sec:imp_details}
\subsection{Architecture and hyper-parameters} 
We implement \algoName using PyTorch 1.4, Python 3.7, and CUDA 10.2.
We train with the Adam optimizer~\cite{kingma2014adam}, with learning rate $ = 4e^{-3}$. 
The \algoName encoder has $L_{e}=4$ layers, and the decoder has $L_d=4$. 
There are $4$ transformer encoder heads, and $4$ decoder heads. 
The effective sequence length (input size of transformer encoder) is $100$.
The representation size of each transformer element is $256$ dimention. For class-balanced focal loss, we set $\beta=0.99$ and $\gamma=2$.

We will make our code publicly available and support academic research purposes. 

\section{Sources of error}
\label{sec:sources_error}
In Sec.~4.4 of the main paper, we presented the different loss functions that comprise our overall objective. Temporal Action Localization involves solving two sub-tasks -- localization and recognition. Hence errors in the final performance can propagate from either of these tasks. 

The first question regarding the overall objective is about the relative weighting of the losses for the two tasks. In Eq.~6 in the main paper, we present the overall objective as simply a sum of the recognition loss and localization loss. In experiments with an earlier \algoName \texttt{w/ GAT} model, we attempt to train the model with larger weights on the classification loss because we observed some room for improvement in recognition. Specifically, we tried weights in the range $[1, 10]$, but this did not improve performance. However, we note that a more thorough hyper-parameter search could indeed result in an improvement in the overall performance of \algoName. 

The other consequence of the multiple loss functions is that the overall performance measured by mAP can be similar for models that achieve different classification and localization loss. 
When we compare Graph Attention \texttt{GAT} and Deformable Attention \texttt{DA} in \algoName, the relative performance improvement with \texttt{DA} appears to be more due to better recognition, than better localization (see \cref{tab:source_error}). 

While there does exist a correlation between task loss and mAP -- larger recognition or localization losses imply lower mAP -- at a fine-grained level, this relationship is imprecise. This is to be expected, given the procedure to calculate mAP. 

\begin{figure}[t]
  \centering
    \includegraphics[width=\linewidth]{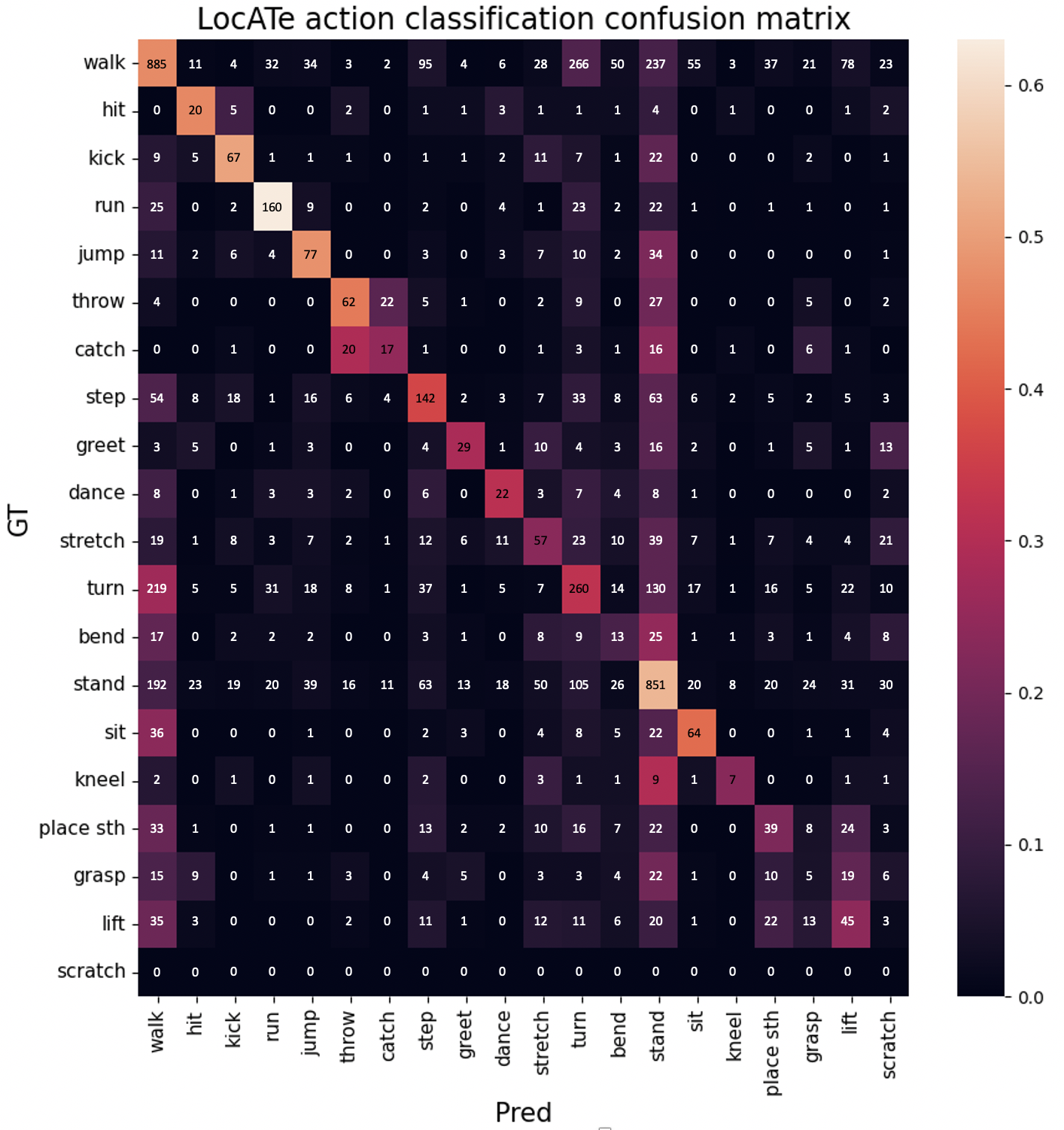}
  \caption{Confusion matrix for action recognition on the predicted spans. 
  Predictions are along x-axis and ground-truth along y-axis. 
  A cell contains the number of samples that is predicted as class $x$, and belongs to class $y$. 
  The background color of a cell indicates the \% of all samples predicted as class $x$ that is actually class $y$.
  Thus, for a cell on the diagonal, it visualizes precision for the class.
  \vspace{-7pt}
  }
  \label{fig:conf_mat}
\end{figure}

\section{Analysis}
\label{sec:analysis}

\noindent
\textbf{Sources of error.} 
In end-to-end training for \tdtal, there are two sources of error -- localization and recognition. 
There often exists a trade-off between the localization loss (Eq. 3) and recognition loss (Eq. 5). 
Though the mAP of two approaches might be identical, a certain architecture performs better localization, while another performs better recognition. 

To understand the recognition error in further detail, in Fig.~\ref{fig:conf_mat}, we visualize the confusion matrix of the action classification on predicted spans. 
The action `stand' is confused with many other actions. 
This is understandable since in the ground-truth data, a person rarely labels `stand' when other actions are present. 
Since we use only $20$ actions, many movements are similar to `stand' but are unlabeled in \btt. 
The actions `throw' and `catch' often occur in frequent succession in movements. 
Interesting that for each action, the other is the most confusing. 
This may be due to the localization error -- if the localization is slightly off from a `catch' in a `throw-catch' sequence, then the ground-truth will be considered to be `throw' (the ground-truth action that corresponds to the predicted span). 

\begin{figure}[t!]
  \centering
    \includegraphics[width=\linewidth]{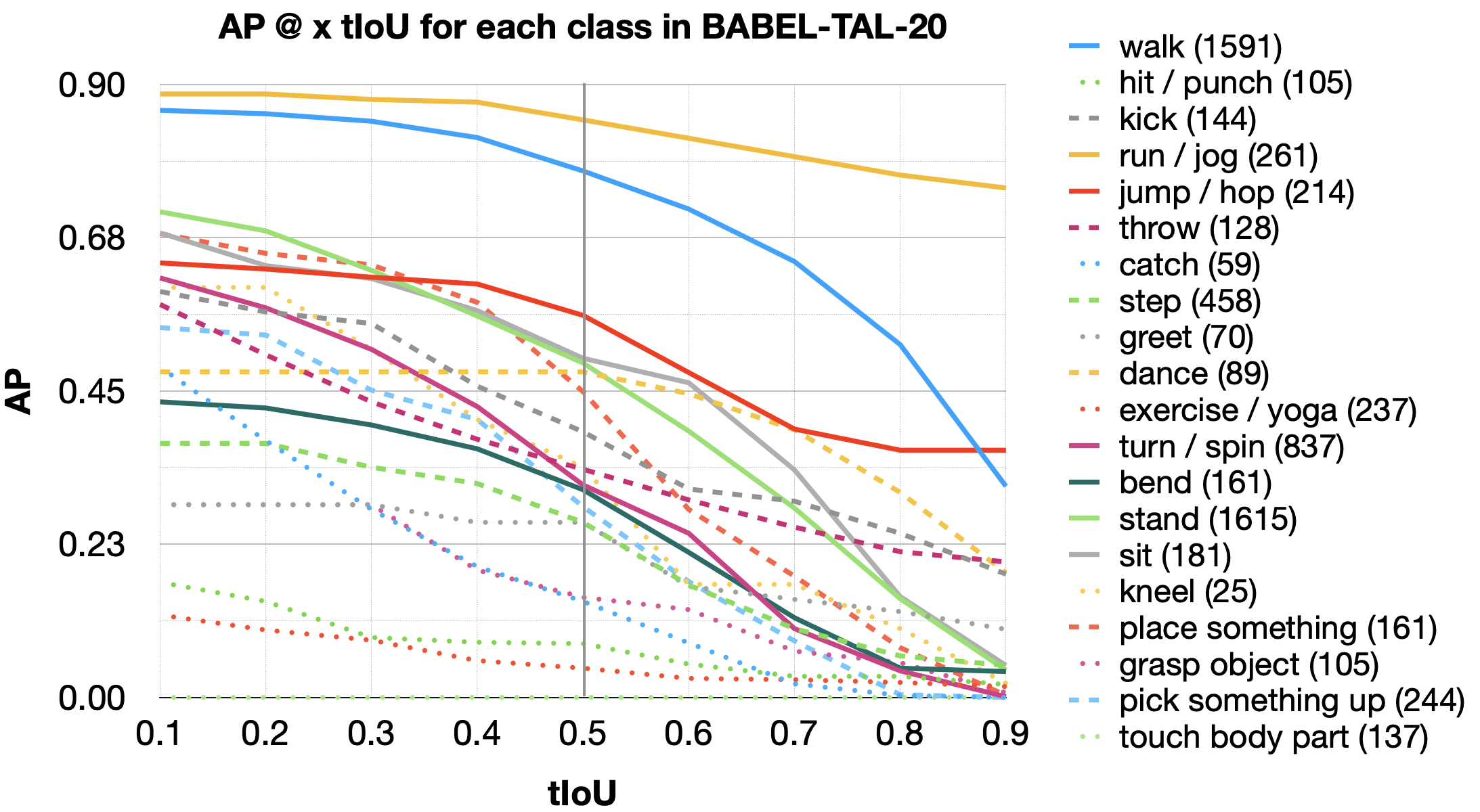}
  \caption{Average Precision (y-axis) for each action in \datasetName (different line styles, colors), across varying tIoU thresholds (x-axis). 
  There exists large variance in the AP of each class; both in absolute magnitude, and in change in AP with tIoU.
  Like any learning method, the performance of \algoName likely improves with the number of samples per class (in parentheses beside actions in legend). However, the absence of a clear correlation between class frequency and AP indicates other contributing factors to the difficulty of a class in \tdtal, e.g., intra-class variance. 
  \vspace{-5pt}
  }
  \label{fig:per_class_AP}
\end{figure}

\noindent \textbf{Performance across actions.} 
We observe from Fig.~\ref{fig:per_class_AP}, that the Average Precision (AP) is highest for the actions `run' and `walk'. While performance on `run' drops gradually with increasing tIoU, the drop is more pronounced for `walk'. 
This indicates that the localization is good for `walk', and extremely good for `run'. 
That is, the overlap between predicted and ground-truth spans is large for `run'. 
There exists large variance in the AP of each class; both in absolute magnitude, and in change in AP with tIoU.
Like any learning method, the performance of \algoName likely improves with the number of samples per class (in parentheses beside actions in legend). However, the absence of a clear correlation between class frequency and AP indicates other contributing factors to the difficulty of a class in \tdtal, e.g., intra-class variance. 
Interestingly, we find that \algoName achieves reasonable AP for `place something', despite not having access to hand positions or finger articulation data. 
Similarly, in Fig.~\ref{fig:conf_mat}, we see that despite the low recall, `grasp something' is in fact, predicted frequently, and often confused with `lift something' and `place something'. 
Grasping an object is often a full-body interaction \cite{GRAB:2020}. It is possible that both the full-body signal, and the temporal context contribute to the inference that the person is grasping or placing something. 

\section{Input features}
\label{sec:inp_feat}

\begin{table}
  \centering
  \caption{Ablation Study: Effect of different 3D Human Representations on \btt. 
  }
  \label{tab:featabl}
  \begin{tabular}{lcc}
\toprule
{\bf Method} & {\bf mAP @ 0.5 tIoU}	\\
\midrule
2D Features \cite{nawhal2021activity} & $14.5$ \\
\textbf{\algoName w/ Graph-attention} \\ 
Joint pos. & $\bm{23.4}$ \\
Early-layer AR feat. & $21.3$ \\
Later-layer AR feat. & $20.4$ \\
Joint pos. + Early + Later AR feat. & $21.4$ \\
    \bottomrule
  \end{tabular}
\end{table}

\subsection{2D Features} 
As a sanity-check, we first ask the following question -- given 3D data, does \tdtal really result in better performance than 2D-TAL on the same data? 
To answer this question, we consider a baseline model \cite{nawhal2021activity} that takes 2D videos as input. 
We utilize the rendered 2D videos of the mocap sequences in \btt as input. 
The human bodies are animated using the SMPL~\cite{SMPL:2015} body model.

We extract I3D~\cite{carreira2017quo} features for the \btt videos, using an I3D model that was pre-trained for activity recognition on real videos from the Kinetics~\cite{carreira2017quo} dataset. 
To obtain I3D features corresponding to an input video with $T$ frames, we first divide the video into short overlapping segments of $8$ frames with an overlap of $4$ frames resulting in $T^{'}$ chunks. 
In other words, we extract features in a sliding-window fashion, with a filter size $=8$ frames and stride $=4$ frames.  
We obtain a tensor of size $T^{'} \times 2048$ as features for these $T^{'}$ chunks. 
Each video feature sequence is rescaled to $100 \times 2048$ (input size of the transformer) using linear interpolation along the temporal dimension. 

\subsection{3D Action Recognition features} 
State-of-the-art 2D-TAL methods \cite{nawhal2021activity} utilize features from a video recognition backbone (e.g. I3D~\cite{carreira2017quo}) as input to the model. 
Compared to raw pixel values from the videos, these features are lower-dimensional, and semantically more meaningful. 
Although using the representation from a feature extractor increases the computation and memory requirements compared to using the `raw' joint positions as input, we attempt to determine if action recognition (AR) features improve performance in \tdtal.

First, we train a popular action recognition model, 2S-AGCN~\cite{shi2019two}, to classify the $20$ actions in the \btt dataset. 
To maximize the discriminativeness of the input feature, we train the recognition model with $8$ frames -- the size of a single input snippet to the \algoName transformer encoder. 
The model achieves a top-1 accuracy of $63.41\%$ and top-5 accuracy of $85.83\%$, demonstrating that it successfully captures some semantic information, despite being far from perfect. 

We then experiment with features extracted from two different layers of the 2S-AGCN model -- after the second graph-convolution layer (Early-layer AR feat.) and after the last graph-convolution layer (Later-layer AR feat.). 

\subsection{Results} 
Table \ref{tab:featabl} shows the results from our experiments with different input representations. 
Unsurprisingly, the 2D features extracted from the rendered videos, under-perform the 3D joint features. This is because the 3D representation contains more information than 2D, which demonstrates the superiority of 3D representation.
We observe that the 3D action recognition features do not improve performance compared to joint position information. 
This implies that 3D input feature representation is still an open problem to explore.
Note that we performed the experiments with \algoName \texttt{w/ GAT}, i.e., \algoName with sparse Graph Attention, which is an earlier model. Since we observed the best performance with joint positions as input, we employed the same in our final model \algoName (with deformable attention \cite{zhu2020deformable}).

\clearpage

%
%
\bibliographystyle{splncs04}
\bibliography{egbib}
\end{document}